\documentclass[letterpaper, 10 pt, conference]{ieeeconf}  

\IEEEoverridecommandlockouts                              
                                                          
\usepackage{graphicx}
\usepackage{cite}

\usepackage{fvextra}
\usepackage{cuted}
\fvset{fontsize=\footnotesize,breaklines=true}
\usepackage{booktabs}
\usepackage{multirow}
\usepackage{amsmath}
\usepackage[table]{xcolor}
\usepackage{eso-pic}

\definecolor{catFunc}{HTML}{0072B2}
\definecolor{catPsych}{HTML}{E69F00}
\definecolor{catMachine}{HTML}{009E73}
\definecolor{catPref}{HTML}{D55E00}
\definecolor{catIdentity}{HTML}{CC79A7}

\usepackage[most]{tcolorbox}
\newtcbox{\catbox}[2][]{on line, arc=3pt, outer arc=3pt, 
  colback=#2!25, colframe=#2!25, 
  boxrule=0pt, boxsep=0pt, left=1pt, right=1pt, top=1pt, bottom=1pt,
  #1}

\newcommand{\catfunc}[1]{\catbox{catFunc}{#1}}
\newcommand{\catpsych}[1]{\catbox{catPsych}{#1}}
\newcommand{\catmachine}[1]{\catbox{catMachine}{#1}}
\newcommand{\catpref}[1]{\catbox{catPref}{#1}}
\newcommand{\catident}[1]{\catbox{catIdentity}{#1}}

\newcommand{\romannoticeoverlay}{%
  \AddToShipoutPictureFG*{%
    \AtPageUpperLeft{%
      \put(\LenToUnit{0.62in},\LenToUnit{-0.52in}){%
        \begin{minipage}{\LenToUnit{7.26in}}%
          \centering\footnotesize\itshape
          Accepted for publication at the 35th IEEE International Conference on Robot and Human Interactive Communication (RO-MAN 2026).%
        \end{minipage}%
      }%
    }%
  }%
}

\title{\LARGE \bf
Why That Robot? A Qualitative Analysis of Justification Strategies for Robot Color Selection Across Occupational Contexts
}

\author{Jiangen He$^{1}$, Wanqi Zhang$^{1}$, and Jessica K. Barfield$^{2}$
\thanks{$^{1}$Jiangen He (ORCID: 0000-0002-3950-6098) and Wanqi Zhang (ORCID: 0009-0007-2376-4942) are with The University of Tennessee, Knoxville, Tennessee, USA ({\tt\small \{jiangen, wzhang79\}@utk.edu})}%
\thanks{$^{2}$Jessica K. Barfield (ORCID: 0000-0002-1208-6118) is with University of Kentucky, Lexington, Kentucky, USA ({\tt\small jessicabarfield@uky.edu})}%
}

\begin{document}

\romannoticeoverlay
\maketitle
\thispagestyle{empty}
\pagestyle{empty}

\begin{abstract}

As robots increasingly enter the workforce, human-robot interaction (HRI) must address how implicit social biases influence user preferences.
This paper investigates how users rationalize their selections of robots varying in skin tone and anthropomorphic features across different occupations.
By qualitatively analyzing 4,146 open-ended justifications from 1,038 participants, we map the reasoning frameworks driving robot color selection across four professional contexts.
We developed and validated a comprehensive, multidimensional coding scheme via human--AI consensus ($\kappa = 0.73$).
Our results demonstrate that while utilitarian \textit{Functionalism} is the dominant justification strategy (52\%), participants systematically adapted these practical rationales that align with existing racial and occupational stereotypes.
Furthermore, we reveal that bias frequently operates beneath conscious rationalization: exposure to racial stereotype primes significantly shifted participants' color choices, yet their spoken justifications remained masked by standard affective or task-related reasoning.
We also found that demographic backgrounds significantly shape justification strategies, and that robot shape strongly modulates color interpretation.
Specifically, as robots become highly anthropomorphic, users increasingly retreat from functional reasoning toward \textit{Machine-Centric} de-racialization.
Through these empirical results, we provide design implications to help reduce the perpetuation of societal biases in future workforce robots.

\end{abstract}

\section{INTRODUCTION}

As robots increasingly enter the workforce across diverse occupational sectors, the potential for human biases to influence robot selection has become a critical topic of discussion within the human-robot interaction (HRI) community.
For HRI, fundamental questions have emerged regarding whether human biases: (i) influence human-robot selection decisions, (ii) shape the strategies users employ when choosing robots for different occupations, and (iii) implicate ethical concerns when users select robots that vary by racial characteristics such as skin tone and anthropomorphic features.

Our previous work examined whether societal biases shape robot selection across construction, healthcare, educational, and athletic domains~\cite{he2026human}, guided by Social Role Theory~\cite{eagly2012social}, which posits that stereotypes emerge from observed group membership across occupational roles, and the Stereotype Content Model~\cite{fiske2018model}, which holds that social perceptions form along two dimensions: warmth and competence.
We found significant biases: healthcare and educational scenarios favored lighter-skinned robots, while construction and athletic contexts showed greater acceptance for darker-toned alternatives.
Moreover, exposure to human professionals from specific racial backgrounds—acting as a stereotype prime—systematically shifted selections in stereotype-consistent directions.
These findings indicate that occupational and color-based biases transfer from inter-human to human-robot evaluation, with significant implications for integrating robots into society.

The companion study ~\cite{he2026human} quantified \textit{what} participants chose but not \textit{why}.
The present study completes the picture by analyzing the open-ended justifications participants provided alongside those same selections, pairing the documented behavioral biases with their own reasoning to move beyond establishing that bias exists toward understanding the rationalization strategies through which it is expressed.
Consequently, the present study employs a qualitative approach, analyzing over 4,000 open-ended responses to examine the reasons behind these selection decisions.
Following an inductive thematic approach, our analysis allows us to gain insights into the human biases and  rationalization strategies that guide robot selection across diverse professional roles.
This approach is motivated by psychological research demonstrating that stereotypes often operate through automatic activation processes, where contextual cues prime specific cognitive schema that influence subsequent judgments~\cite{schneider2019application}.
By using qualitative methods, we explore the nuanced rationales participants provide for their selections.

Further, addressing our topic allows us to focus on an important and emerging issue for the HRI community, that is, issues of ethics for AI-guided robots in an age of increasingly smart robots which display social identities~\cite{barfield2026evaluating}.
Our current paper allows us to address a gap in the HRI literature regarding the understanding of reasons and strategies that are used when robots are selected to enter different occupational contexts.
To add to the emerging stream of research on human biases in robotics, in the current study we used open-ended questions to elicit the reasons why participants made selections in different contexts.
Using a systematic qualitative coding process, we first created a scheme to identify the reasons why users selected robots for different occupations, we next evaluated the reliability of the coding scheme, and concluded with design guidelines for the selection of robots entering the workforce.
We organize our findings around four research questions:

\begin{itemize}
    \item \textbf{RQ1}: What reasoning frameworks do participants employ to justify robot color selection?
    \item \textbf{RQ2}: How do situational factors---task context, robot color, and racial priming---shape justification patterns?
    \item \textbf{RQ3}: How do participant demographics interact with robot color to shape justification strategies?
    \item \textbf{RQ4}: How does robot human-likeness modulate the relationship between appearance and justification?
\end{itemize}

\section{Related Work}

\subsection{Social Categorization and Robot Racialization}

Emerging literature demonstrates that humans project social categories onto robots.
Research has shown that humans display racial bias toward robots in shooter-bias tasks~\cite{addison2019robots,bartneck2018robots}, and robot skin tone interacts with user prejudice to shape attributions of agency~\cite{eyssel2013don}.
Sparrow argues that race, as a social construction, inevitably extends to robots, carrying significant ethical implications~\cite{sparrow2019robots}.
These studies indicate that HRI is not immune to social categorization, providing a foundation for examining how users perceive and categorize robots based on visual cues.

\subsection{Occupational Contexts and Role-Based Stereotypes}

Occupational roles are potent sources of stereotype activation.
Social role theory argues that stereotypes emerge from observed occupational distributions~\cite{koenig2014evidence}, while the Stereotype Content Model (SCM) shows systematic associations between social categories and occupational roles along competence and warmth dimensions~\cite{cuddy2009stereotype,oliveira2019stereotype}.
Empirical evidence reveals stereotypical linkages: Latinos with manual labor, Asians with academic competence, Blacks with athletic ability, and Whites with professional roles~\cite{oliveira2019stereotype,koenig2014evidence,pager2008sociology}.
In HRI, such contexts may prime users to expect certain robotic appearances that align with these established human stereotypes.

\subsection{Anthropomorphism and Human-Likeness}
Anthropomorphism determines how strongly social categories influence user perceptions.
While anthropomorphic design can increase trust, it also amplifies the salience of social category cues, such as skin tone~\cite{cheetham2011human,lee2016humanlike}.
Physical human-likeness interacts with warmth and competence cues to shape robot evaluations~\cite{belanche2021examining}.
Specifically, anthropomorphic features can trigger deeper social judgments, making robots more vulnerable to stereotype-based evaluations when occupational priming is present~\cite{eyssel2013don,cumbal2023stereotypical}.

\subsection {Interpreting Appearance Cues in Human–Robot Interaction}
Beyond social categorization, robot appearance also shapes how users interpret a robot's role and capabilities.
Visual features often function as heuristic cues that people use to infer the intended purpose of technological artifacts~\cite{krippendorff2005semantic}, so appearance cues such as form, color, and material can influence perceptions of competence, professionalism, and task suitability.
For example, darker colors are often associated with durability and strength, whereas lighter colors are linked to cleanliness and approachability in service contexts~\cite{elliot2015color,https://doi.org/10.1002/mar.21268}.
Users may thus interpret robot appearance not only through social categories but also through contextual expectations about how a robot should look in a given occupational setting.

\section{Methods}

\subsection{Study Design and Participants}

This paper analyzes the open-ended justifications collected in the same two experiments whose quantitative selection results are reported in our companion study~\cite{he2026human}; the participants, stimuli, and procedures are therefore identical, and we summarize the design here to make the present analysis self-contained.
Study~1 employed a $4 \times 5$ mixed factorial design ($N = 421$): task scenario (four professional contexts) was a within-subjects factor, and human-likeness level (five levels) was a between-subjects factor.
On each trial, participants read a short written scenario and viewed a matching picture of the task setting that did not depict people, isolating baseline contextual associations without racial priming.
They then selected one robot from the six color options at their assigned human-likeness level and provided an open-ended written justification for that choice.

Study~2 added a racial priming manipulation, employing a $2 \times 4 \times 5$ mixed factorial design ($N = 617$).
Task scenario and priming condition (stereotype vs.\ non-stereotype) were within-subjects factors: every participant completed all four tasks, two under \textit{Stereotype priming} and two under \textit{Non-stereotype priming}, with random assignment of priming type to task.
Human-likeness level was a between-subjects factor.
The priming manipulation embedded human professional imagery in the task scenario picture, investigating how racial representations shape robot selection decisions, which simulated how the inter-human bias transfer to human-robot interaction.
\textit{Stereotype priming} paired each task with a human professional from a racial background commonly associated with that occupation in popular representations (construction: Latino; hospital: White; tutoring: Asian; sports: Black).
\textit{Non-stereotype priming} used less typical pairings for the same tasks (e.g., White construction, Black hospital, Latino tutoring, Asian sports).

\begin{figure}[t]
\centering
\includegraphics[width=0.8\linewidth]{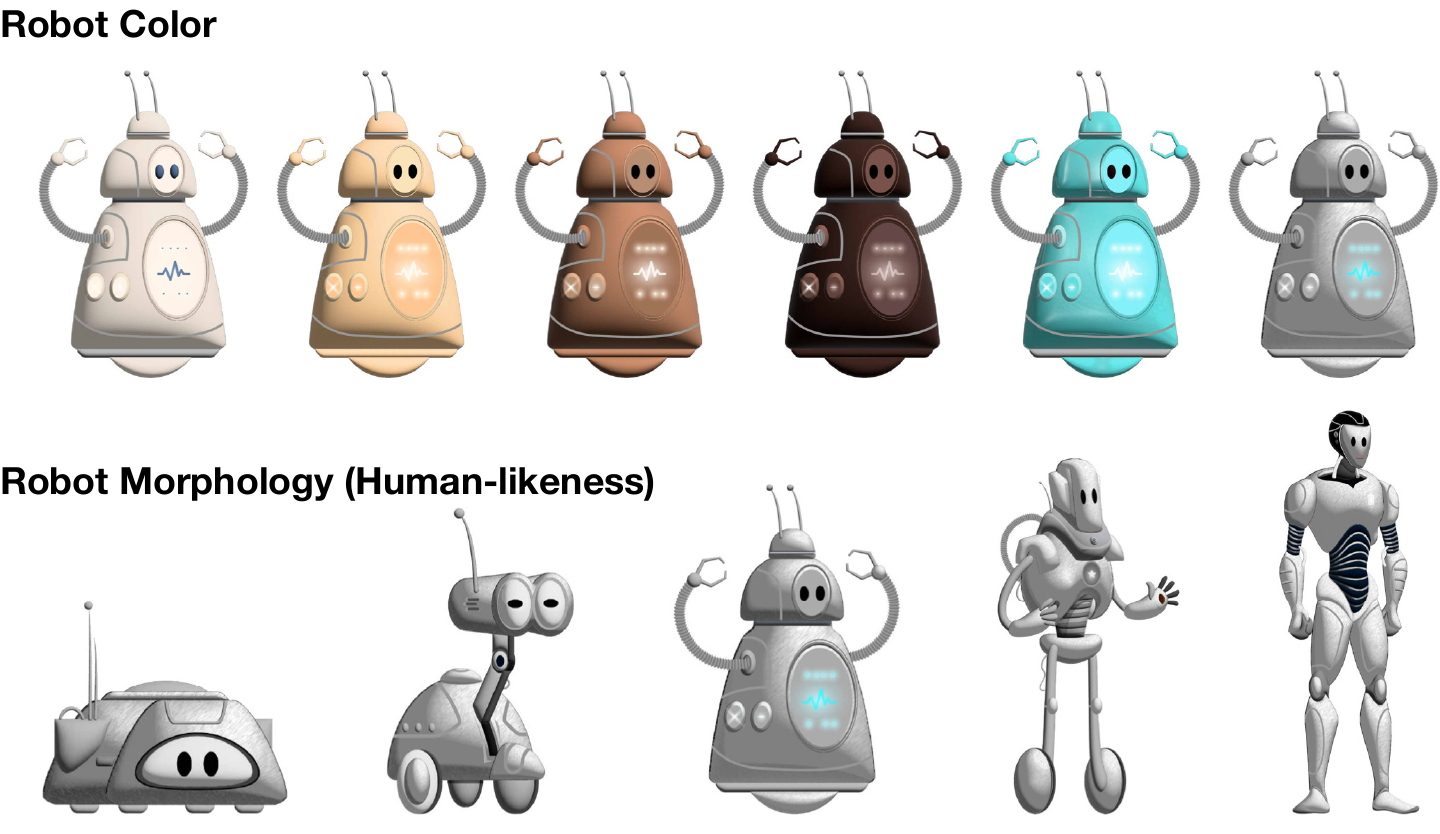}
\caption{Representative robot stimuli illustrating the six color options (columns) and five human-likeness levels (rows).
Skin-tone colors are Light, Medium, Brown, and Dark; Silver and Teal serve as non-skin-tone baselines.}
\label{fig:robots_stimuli}
\end{figure}

Robot stimuli were rendered illustrations varying in five human-likeness levels and six color options.
Figure~\ref{fig:robots_stimuli} shows representative examples; the experimental workflow and complete stimulus set appear in the Appendix.

Participants were recruited from the United States through Prolific, with quota sampling targeting balanced distributions by gender and race.
In both studies, participants selected one of six robots varying in skin tone (Light, Medium, Brown, Dark) and non-skin-tone colors (Silver, Teal) within their assigned human-likeness level, across four professional task scenarios: construction site, hospital, home tutoring, and sports field.
After each selection, participants provided an open-ended explanation for their choices ($N_{participants} = 1,038,N_{repsonses}=4,146$).

\subsection{Codebook Development}

To systematically analyze the open-ended justifications, we developed a multidimensional coding scheme using an inductive thematic approach.
The process followed three iterative phases.

\paragraph{Phase 1: Stratified Sampling.}
From the 4,146 responses, we constructed a representative training subset of $N = 500$ ($n = 250$ per study).
We employed stratified sampling based on task scenario, robot color, and human-likeness to ensure that the codebook captured the full breadth of reasoning, including minority choices and diverse occupational contexts.

\paragraph{Phase 2: Open Coding and Constant Comparison.}
The sampled responses underwent line-by-line open coding to identify recurring concepts and descriptive labels (e.g., ``dirt concealment,'' ``professionalism'').
Following the Constant Comparative Method \cite{rankin2017comparative}, we iteratively compared new data incidents with existing codes to refine definitions and establish conceptual boundaries.

\paragraph{Phase 3: Axial Coding and Saturation.}
In the final stage, these codes were synthesized through axial coding into higher-level thematic clusters representing distinct reasoning strategies.
This iterative process continued until theoretical saturation was achieved—the point at which no new categories emerged from the data.
The finalized codebook, with precise definitions and anchor examples, then guided single-label coding of the full dataset, assigning each response exactly one main category and one sub-category that best captured its primary reasoning.
The codebook captures a spectrum from practical reasoning (C1: Functionalism) through affective responses (C2: Psych/Affective) to explicit social mapping (C5: Identity/Social).
C3 (Machine-Centric) captures a distinctive response pattern in which participants actively resist the humanization of robots, while C4 (Preference/Evasion) captures low-effort or avoidant responses.
Table~\ref{tab:codebook} presents the five main categories and fifteen sub-categories.

\begin{table*}[t]
\caption{Codebook: Five main categories and fifteen sub-categories for robot color selection justifications.}
\label{tab:codebook}
\footnotesize
\begin{tabular}{@{}p{0.13\textwidth}p{0.21\textwidth}p{0.34\textwidth}p{0.24\textwidth}@{}}
\toprule
\textbf{Main Category} & \textbf{Sub-Category} & \textbf{Definition} & \textbf{Example Keywords} \\
\midrule
\multirow{5}{*}{\parbox{0.16\textwidth}{C1. Functionalism \&\\Task-Fit}}
 & C1.1 Durability \& Ruggedness & Physical strength or resistance to damage & ``tough,'' ``strong,'' ``sturdy'' \\
 & C1.2 Cleanliness \& Maintenance & Color relates to environment's cleanliness standards & ``sterile,'' ``hide dirt,'' ``hygiene'' \\
 & C1.3 Visibility \& Safety & Robot needs to be seen for safety or tracking & ``stand out,'' ``high visibility'' \\
 & C1.4 Professionalism \& Fit & Matching industry norms or environmental aesthetics & ``scrubs,'' ``professional'' \\
 & C1.5 Performance/ Capability & General attribution of skill based on appearance & ``capable,'' ``efficient,'' ``athletic'' \\
\midrule
\multirow{3}{*}{\parbox{0.16\textwidth}{C2. Psychological \&\\Affective Impact}}
 & C2.1 Approachability \& Trust & Perceived as non-threatening or welcoming & ``friendly,'' ``calm,'' ``soothing'' \\
 & C2.2 Engagement \& Animation & Bright colors will motivate or maintain attention & ``engaging,'' ``fun,'' ``vibrant'' \\
 & C2.3 Human-Likeness (Social) & Chosen because it looks ``more human'' & ``relatable,'' ``looks like a person'' \\
\midrule
\multirow{3}{*}{\parbox{0.16\textwidth}{C3. Machine-Centric\\Heuristics}}
 & C3.1 Materiality \& Industrialism & Associating color with metal, steel, or machinery & ``looks like metal,'' ``utilitarian'' \\
 & C3.2 Neutrality & Robots should not have human skin tones & ``neutral,'' ``shouldn't have race'' \\
 & C3.3 Anti-Distraction & Color chosen to minimize attention to the robot & ``non-distracting,'' ``unnoticed'' \\
\midrule
\multirow{2}{*}{\parbox{0.16\textwidth}{C4. Preference \&\\Evasion}}
 & C4.1 Personal Aesthetic & Simple statement of personal color preference & ``I like the color,'' ``my favorite'' \\
 & C4.2 Random/Equivalence & Admitting choice was arbitrary & ``randomly picked,'' ``no reason'' \\
\midrule
\multirow{2}{*}{\parbox{0.16\textwidth}{C5. Identity \&\\Social Mapping}}
 & C5.1 Stereotype Alignment & Color matches expected demographic of a profession & ``looks like an athlete,'' ``boss color'' \\
 & C5.2 Identity Projection & Robot mirrors participant's own characteristics & ``reflection of myself'' \\
\bottomrule
\end{tabular}
\end{table*}

\subsection{Inter-rater Reliability Analysis}
We stratified sampled 100 responses for the reliability test, exclusive from the 500 responses for codebook development. Two coders coded the 100 responses separately. The reliability at the main category level yielded Cohen's $\kappa$ values of 0.66 (Coder~1 vs.\ Coder~2, 78\% agreement). At the sub-category level, reliability was slightly lower: $\kappa = 0.60$ (Coder~1 vs.\ Coder~2, 64\% agreement). To establish a definitive human baseline, discrepancies between the two human coders were resolved by a third independent coder. The coders are the authors of this paper.

\subsection{AI-Assisted Coding and Validation}

Given the scale of the dataset ($N = 4{,}146$) and the substantial agreement between two human coders (0.66 and 0.60), we employed an AI-assisted coding approach using Google's Gemini API (model \texttt{gemini-3.1-pro-preview} in our implementation).
The system instruction, including the embedded codebook, appears in Appendix.
Each response was supplied in a separate user message, \texttt{Participant's reason for choosing the robot: ``\ldots''}, and the model returned structured JSON with exactly one main category and one sub-category and was set \texttt{temperature=0}.

We validated the AI coding through inter-rater reliability analysis on the same 100 responses, coded independently by three human coders and the AI system. First, we compared AI with results from Coder 1 and Coder 2. At the main category level, the reliability yielded Cohen's $\kappa$ values of 0.61 (Coder~1 vs.\ AI, 75\% agreement), and 0.73 (Coder~2 vs.\ AI, 83\% agreement).
At the sub-category level, results were 0.60 (Coder~1 vs.\ AI, 64\% agreement) and 0.72 (Coder~2 vs.\ AI, 75\% agreement). Second, comparing this reconciled human consensus against the AI output demonstrated strong alignment, yielding $\kappa = 0.73$ (83\% agreement) at the main category level and $\kappa = 0.69$ (72\% agreement) at the sub-category level.
Based on standard interpretive criteria~\cite{landis1977measurement}, these values reflect substantial agreement, demonstrating that the AI system's categorization reliably mirrors human judgment.

\subsection{Analysis}

We employed a mixed-methods approach to analyze the coded justifications.
The quantitative component computed frequency distributions and tested associations using Pearson's $\chi^2$ tests of independence, with Cram\'{e}r's $V$ as the effect size measure.
The qualitative component added targeted thematic analyses of sub-categories of particular interest, examining their distributions across demographics, task contexts, and robot characteristics and supplementing the statistical findings with illustrative participant quotes.
This dual approach maps broad statistical patterns while preserving the nuanced contexts of user rationalizations.

\section{Results}

\subsection{RQ1: The Landscape of Justification}

\begin{table}[t]
\centering
\caption{Summary of justification categories.
Skin-tone \% indicates the proportion of responses within each category where a skin-tone robot was selected.}
\label{tab:summary}
\footnotesize
\begin{tabular}{@{}lrrrr@{}}
\toprule
\textbf{Category} & \textbf{N} & \textbf{\%} & \textbf{Skin\%} & \textbf{Words} \\
\midrule
 Functionalism       & 2,163 & 52.2 & 49.6 & 14.4 \\
 Psych/Affective     &   912 & 22.0 & 49.8 & 14.7 \\
 Preference/Evasion  &   602 & 14.5 & 50.0 & 10.9 \\
 Machine-Centric     &   421 & 10.2 & 27.1 & 16.0 \\
 Identity/Social     &    48 &  1.2 & 85.4 & 15.1 \\
\midrule
\textbf{Total}      & 4,146 & 100  & 47.7 & 14.0 \\
\bottomrule
\end{tabular}
\end{table}

\begin{figure}[t]
\centering
\includegraphics[width=\linewidth]{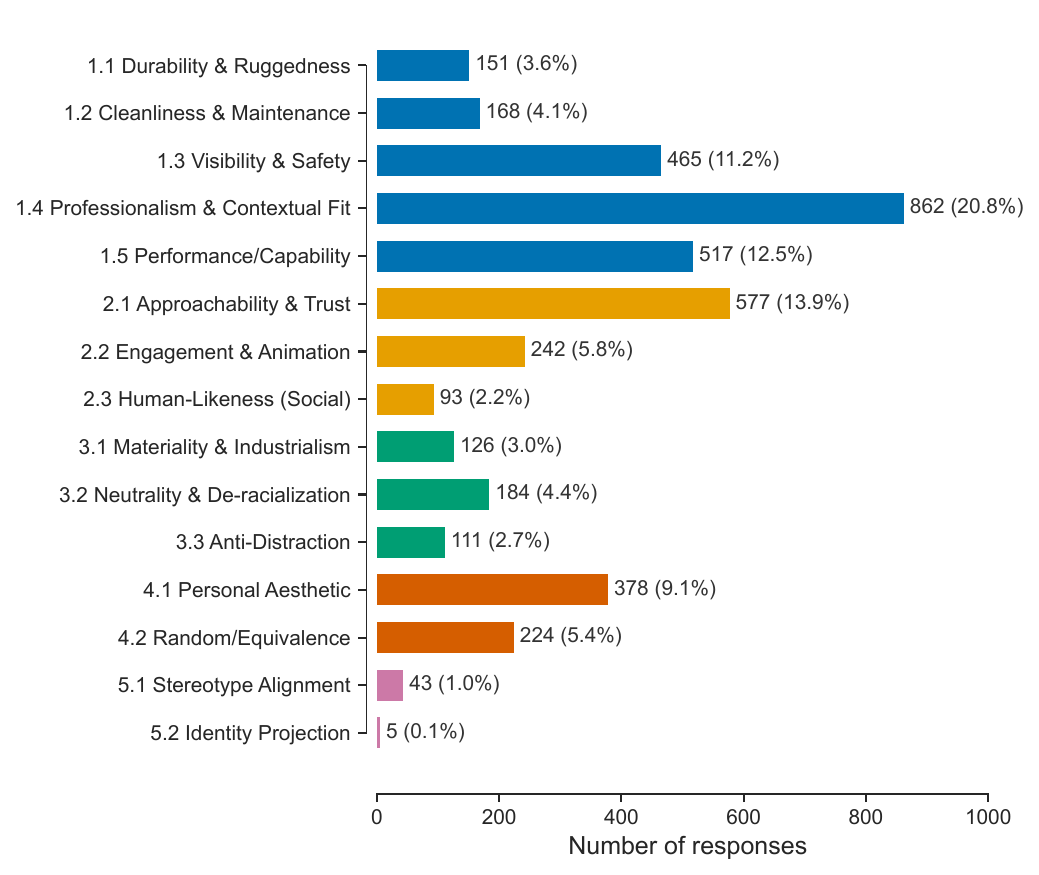}
\caption{Frequency distribution of the fifteen sub-categories.}
\label{fig:sub_category_freq}
\end{figure}

The distribution of reasoning frameworks across all 4,146 coded responses reveals a clear hierarchy (Table~\ref{tab:summary}; see the Appendix for detailed category explanations and representative quotes).
\catfunc{Functionalism} dominated, accounting for over half of all justifications (52.2\%).
\catpsych{Psych/Affective} was the second most common framework (22.0\%), followed by \catpref{Preference/Evasion} (14.5\%) and \catmachine{Machine-Centric} (10.2\%).
\catident{Identity/Social} comprised only 1.2\% of responses.
\catmachine{Machine-Centric} responses skewed toward non-skin-tone choices (27.1\% skin-tone) and \catident{Identity/Social} responses skewed toward skin-tone robots (85.4\%).
Figure~\ref{fig:sub_category_freq} presents the frequency distribution at the sub-category level.

\subsection{RQ2: Situational Factors Shape the Justification}
\subsubsection{Task Context and Justification}

\begin{figure}[t]
\centering
\includegraphics[width=0.8\linewidth]{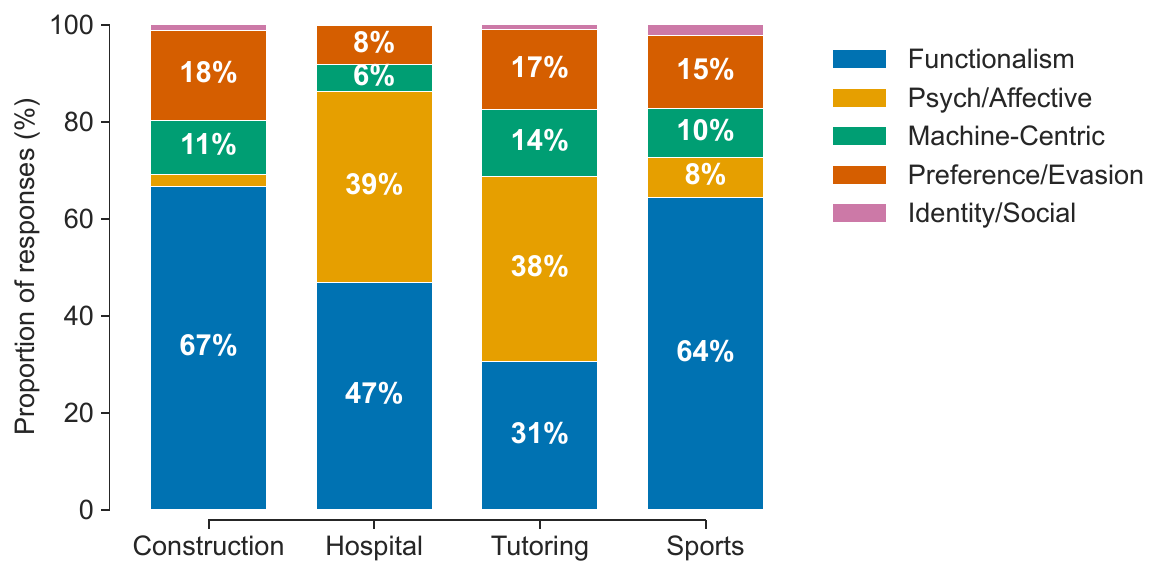}
\caption{Proportion of justification categories across the four task contexts.}
\label{fig:category_by_task}
\end{figure}

\begin{figure}[t]
\centering
\includegraphics[width=0.9\linewidth]{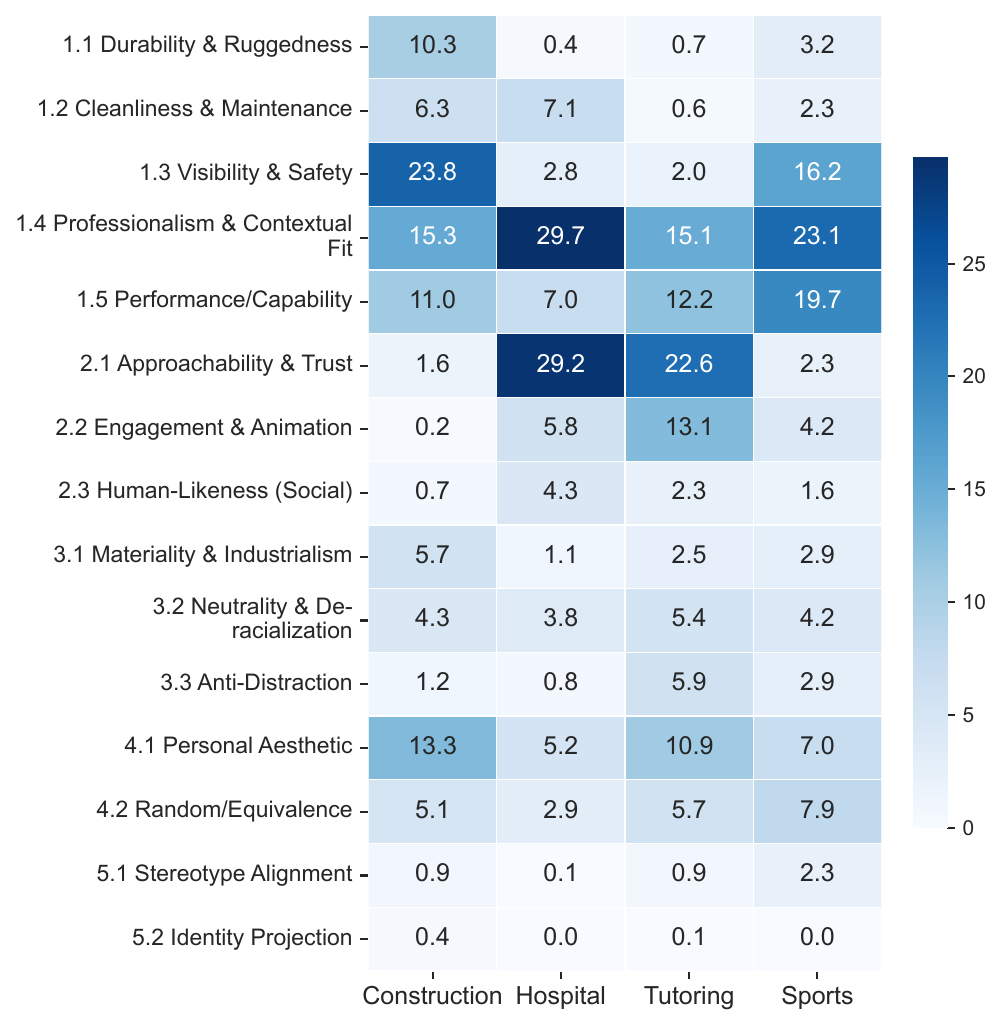}
\caption{Sub-category proportions (\%) within each task context.
}
\label{fig:subcategory_task_heatmap}
\end{figure}

Justification patterns varied significantly across the four professional task contexts ($p < .001$, Cram\'{e}r's $V = 0.25$), representing a medium-to-large effect (Figure~\ref{fig:category_by_task}).
At the sub-category level, this contextual dependence was even stronger ($p < .001$, $V = 0.35$).
Figure~\ref{fig:subcategory_task_heatmap} provides a granular view of these sub-category distributions across the four task contexts.

The construction scenario elicited the highest proportion of \catfunc{Functionalism} responses (68\%), driven by \catfunc{Visibility} (24.5\%) and \catfunc{Professionalism} (16.5\%), as well as the highest rates of \catfunc{Durability} (10.1\%) across all tasks.
Participants selecting darker robots for construction frequently invoked ruggedness: ``\textit{A construction site is likely dirty, so the brown color will show less dirt}'' (White participant, Dark robot).
While the sports scenario shared a similar overall dominance of \catfunc{Functionalism} responses with construction, the internal distribution of sub-categories was slightly different, emphasizing \catfunc{Professionalism} (23.1\%) and \catfunc{Performance} (19.7\%).
Sports also elicited the most frequent explicit \catident{Stereotype Alignment} responses (1.9\%).

The hospital context showed a distinctive surge in \catpsych{Approachability} (29.7\%) and \catfunc{Professionalism} (29.0\%), and elicited the highest rate of \catfunc{Cleanliness} (6.8\%) across all tasks.
Robots in hospitals were justified through sterility associations: ``\textit{White color looks cleaner}'' (White participant, Light robot); ``\textit{Hospitals should be clean so a lot of the features are white so any contamination can be spotted}'' (East Asian participant, Light robot).
The tutoring scenario produced the most evenly distributed pattern across sub-categories.
While \catpsych{Approachability} (22.7\%) remained high, tutoring uniquely emphasized \catpsych{Engagement} (12.7\%---more than double the rate of any other task) and \catmachine{Anti-Distraction} (5.9\%).

\subsubsection{Robot Color Choice and Justification.}
Justification categories also varied significantly by robot color chosen ($p < .001$, $V = 0.19$).
When the analysis was simplified to skin-tone versus non-skin-tone robots, the association remained highly significant ($p < .001$, $V = 0.16$), driven primarily by the concentration of \catmachine{Machine-Centric} reasoning among non-skin-tone selections.

\begin{figure}[t]
\centering
\includegraphics[width=0.9
\linewidth]{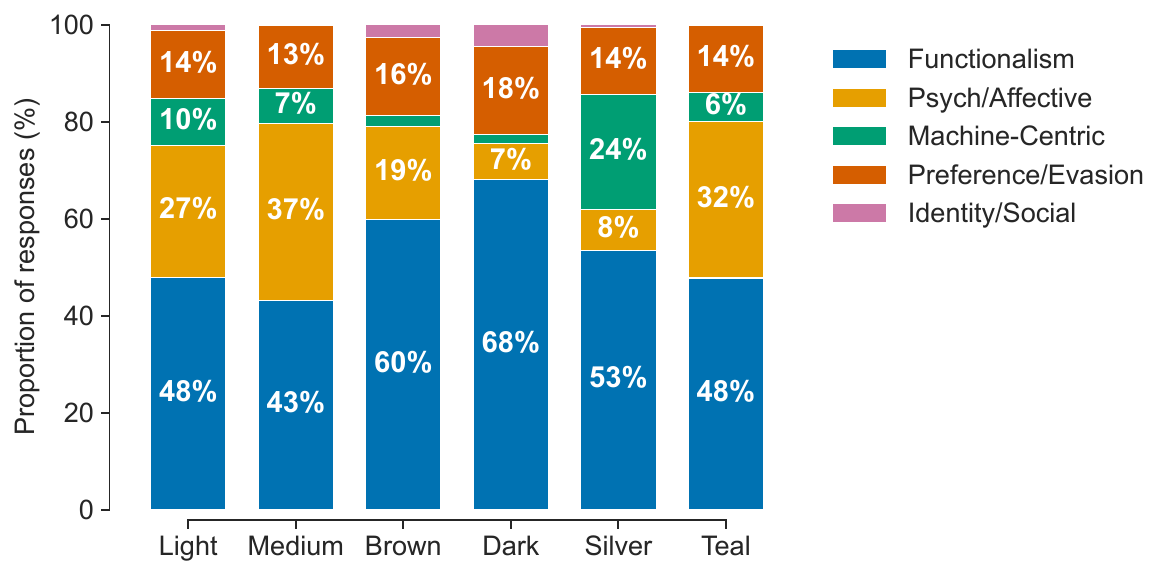}
\caption{Proportion of justification categories by robot color choice. 
}
\label{fig:category_by_color}
\end{figure}

Figure~\ref{fig:category_by_color} illustrates these distributional differences across the six robot colors.
A clear pattern emerges along the skin-tone spectrum: lighter-colored robots were disproportionately justified through \catpsych{Psych/Affective} reasoning, whereas darker-colored robots were more frequently justified through \catfunc{Functionalism}.
Notably, Brown and Dark robots attracted higher rates of functional justifications than even the two non-skin-tone alternatives (Silver and Teal), suggesting that darker skin tones activated task-oriented rationalizations more than non-skin-tone appearance.
\catmachine{Machine-Centric} reasoning peaked sharply among Silver selections, reflecting participants' tendency to invoke materiality when selecting robots with metallic appearance.



\subsubsection{Priming Shifts Choices}

Although racial priming through human professional imagery significantly shifted robot color selections in the companion study~\cite{he2026human}, the presence of these primes only weakly altered the overall distribution of justification categories relative to the unprimed baseline, at both the category ($p = .001$, $V = 0.07$) and sub-category ($p = .001$, $V = 0.09$) levels (see Appendix).
\begin{figure*}[t]
\centering
\includegraphics[width=0.9\textwidth]{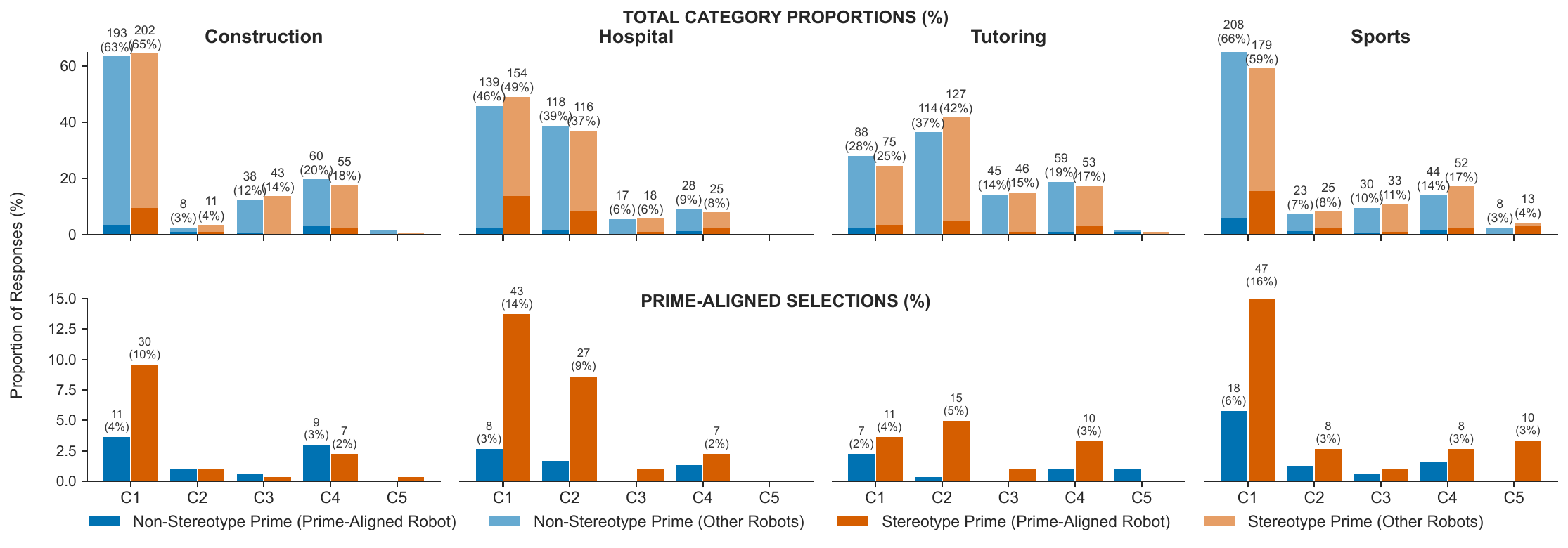}
\caption{Interaction between priming condition, justification category, and prime-aligned robot selection. 
The Top Row displays the overall proportion of each justification category under Stereotype and Non-Stereotype priming conditions. 
Each bar is subdivided: the darker (bottom/base) segment represents choices where the selected robot color aligned with the racial prime (e.g., selecting a Dark robot after a Black prime), while the lighter (top) segment represents all other color selections.
The Bottom Row isolates only the prime-aligned selections, showing the darker segments from the top row. 
Categories: \catfunc{C1: Functionalism}, \catpsych{C2: Psych/Affective}, \catmachine{C3: Machine-Centric}, \catpref{C4: Preference/Evasion}, \catident{C5: Identity/Social}.}
\label{fig:category_by_stereotype_priming}
\end{figure*}


To probe whether the types of racial prime influenced reasoning, we compared \textit{stereotype priming} versus \textit{non-stereotype priming} trials in the primed subset from Study~2 ($N = 2{,}464$).
As Figure~\ref{fig:category_by_stereotype_priming} (top row) illustrates, the overall category distributions are similar between the two conditions across all four tasks. The colors actually selected, however, differed markedly for \textit{prime-aligned} choices---those where the selected robot color matched the skin tone associated with the primed racial group (Latino$\to$Brown, Black$\to$Dark, White$\to$Light, Asian$\to$Medium).
Under stereotype-consistent priming, 19.2\% of selections were prime-aligned, compared with only 6.9\% under stereotype-inconsistent priming.

As Figure~\ref{fig:category_by_stereotype_priming} (bottom row) shows, prime-aligned choices increased sharply under stereotype-consistent priming across every task, consistent with the companion study~\cite{he2026human}.
The justifications underlying these aligned choices, however, adapted specifically to the task context, as the following qualitative examination reveals.

\paragraph{Construction}
In the construction scenario, stereotype priming (Latino) generated more prime-aligned selections (Brown robots) than non-stereotype priming (White $\to$ Light robots).
Both groups relied heavily on \catfunc{Functionalism}, but their functional rationales adapted to the color chosen.
Participants in the non-stereotype condition justified their Light robot selections through general performance metrics: \textit{``The robot looks strong and durable''} (White prime, Light robot).
In contrast, participants selecting Brown robots under stereotype priming seamlessly linked the darker color to the reality of construction work: \textit{``With the light brown color it won't look as dirty when covered in dirt and dust...''} or invoked visibility norms: \textit{``Orange is something workers will be able to easily identify on site''} (Latino prime, Brown robot).
Stereotype priming thus shifted the functional criteria from abstract capability to material resilience and visibility.

\paragraph{Hospital}
In the hospital context, aligned selections were overwhelmingly driven by \catfunc{Functionalism} and \catpsych{Psych/Affective}.
Under stereotype-consistent priming (White), participants actively linked their Light robot selections to professional aesthetics (\textit{``matches the vibe of a hospital,''} \textit{``mimics the white coat''}) and uniquely invoked \catfunc{Cleanliness \& Maintenance} (\textit{``looks more sterile,''} \textit{``represents cleanliness''}).
They also leveraged \catpsych{Psych/Affective} to justify the color as \textit{``pleasant and friendly''} or \textit{``somewhat calming to look at.''}
In contrast, non-stereotype aligned choices (e.g., Latino $\to$ Brown) were not only far less frequent but also relied on much vaguer functional justifications (\textit{``it seems as though it can get the job done,''} \textit{``I would like it be a hospital robot''}) .
This indicates that the stereotype-consistent ``white=sterile/calm'' association was uniquely potent in healthcare settings.

\paragraph{Tutoring}
For the tutoring scenario, stereotype priming (Asian) increased selections of the Medium robot.
The stereotype was manifested through two distinct pathways: \catpsych{Psych/Affective} and \catpref{Preference/Evasion}.
Aligned \catpsych{Psych/Affective} responses mapped the Asian stereotype's dimensions of competence and warmth onto the robot's appearance, describing the Medium robot as \textit{``calm, friendly, and smart for teaching''} or \textit{``warm... yet serious enough to make sure lessons are being taught.''}
\catpref{Preference/Evasion} responses were frequent for this aligned choice: participants claimed they \textit{``selected beige at random,''} or insisted \textit{``the color has nothing to do with tutoring.''}
The Asian prime successfully drove participants to select the stereotype-congruent robot (Medium), but because the stereotype is cognitive rather than physical, participants could not easily construct a task-based functional rationale. 

\paragraph{Sports}
The sports context elicited the most direct expression of stereotype alignment.
Under stereotype priming (Black), Dark robot selections surged.
These choices were justified through \catfunc{Functionalism} focusing intensely on physical dimensions: \textit{``Looks very athletic and strong,''} \textit{``more bold daunting trainer feel.''}
Furthermore, sports was the only context where \catident{Identity/Social} appeared prominently in aligned choices.
Participants explicitly voiced the stereotype that the prime had activated: \textit{``The color shows masculinity''} and \textit{``because most athletes... are black.''}
The stereotype was not merely rationalized via functional heuristics; it was consciously acknowledged as the basis for the decision.



\subsection{RQ3: Participant Demographics and Robot Color Interaction}

\subsubsection{Participant Races}
Justification strategies varied significantly by participant race ($p < .001$, $V = 0.10$; see Appendix), though with a small effect size.
For instance, Black/African American participants exhibited the highest rates of \catfunc{Functionalism} reasoning (59.6\% of their responses, compared to 47.3\%--50.8\% for other groups) and the lowest rates of \catmachine{Machine-Centric} reasoning (4.8\% vs.\ 8.2\%--16.5\%).
Conversely, White participants were the most likely to employ \catpref{Preference/Evasion} heuristics (18.6\% vs.\ 11.1\%--15.4\%).
At the sub-category level, this association was more pronounced ($p < .001$, $V = 0.15$; see Appendix).
The high rate of \catfunc{Functionalism} among Black/African American participants was driven heavily by \catfunc{Performance/Capability (C1.5)}, which they used in 19.9\% of responses compared to only 5.4\% for Asian participants.
White participants' elevated Preference reasoning was largely due to simple \catpref{Personal Aesthetic (C4.1)} justifications (11.1\%, compared to 7.1\% for Asian participants).

\subsubsection{Interaction of Race, Robot Color, and Justification}
\begin{figure}[t]
\centering
\includegraphics[width=0.9\linewidth]{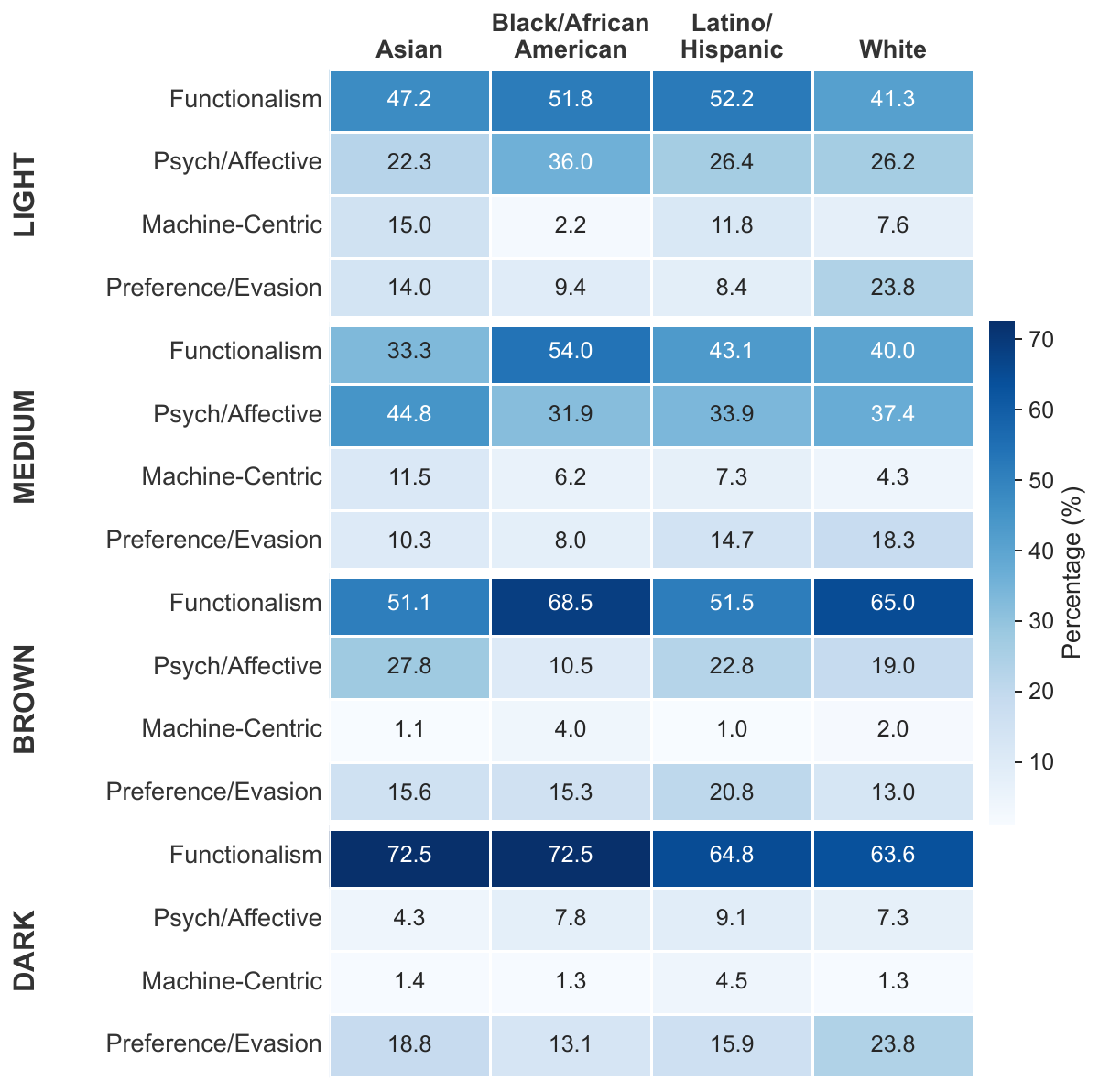}
\caption{The distribution of justification categories by participant race for each selected robot color.
Note: \catident{Identity/Social} is excluded.}
\label{fig:triple_race_cat_color}
\end{figure}

To understand how demographic factors interact with robot color, we examined the interaction between participant race, robot color, and justification category (Figure~\ref{fig:triple_race_cat_color}).

Figure~\ref{fig:triple_race_cat_color} reveals two key patterns.
First, \catfunc{Functionalism} dominates the justification landscape for darker-colored robots across all racial groups.
Second, it reveals race-specific elevations in \catpsych{Psych/Affective} and \catpref{Preference/Evasion} reasoning for robots whose color approximates each group's own skin tone.
Asian participants justified Medium-robot selections through \catpsych{Psych/Affective} reasoning at a distinctly elevated rate (46.4\%) compared to their rates for other colors and compared to other racial groups selecting Medium robots (Black: 31.9\%, Latino: 35.1\%, White: 37.0\%).
Similarly, Latino/Hispanic participants showed elevated \catpref{Preference/Evasion} reasoning for Brown robots (22.5\%) and lower \catfunc{Functionalism} (48.6\%) compared to other groups selecting Brown.
White participants disproportionately invoked both \catpsych{Psych/Affective} (24.6\%) and \catpref{Preference/Evasion} (23.3\%) reasoning for Light robots, with correspondingly lower \catfunc{Functionalism} (43.2\%).
However, Black/African American participants did not exhibit an analogous affective or preferential elevation for Dark robots; instead, their Dark-robot justifications were overwhelmingly functional (75.2\%), mirroring the general \catfunc{Functionalism} trend rather than reflecting any identity-congruent affective pull. 

This asymmetry suggests that the mechanisms linking participant identity to robot color selection operate differently across racial groups, offering a nuanced perspective on racial mirroring---the phenomenon where users project trust or affective resonance onto agents that reflect their own racial identity~\cite{liao2020racial}.
The affective pull of a color-congruent robot is therefore not universal; instead, mirroring is mediated by the specific intersection of racial identity with the robot's appearance.

\subsection{RQ4: Robot Human-Likeness Modulates Justification}
\begin{figure}[t]
\centering
\includegraphics[width=\linewidth]{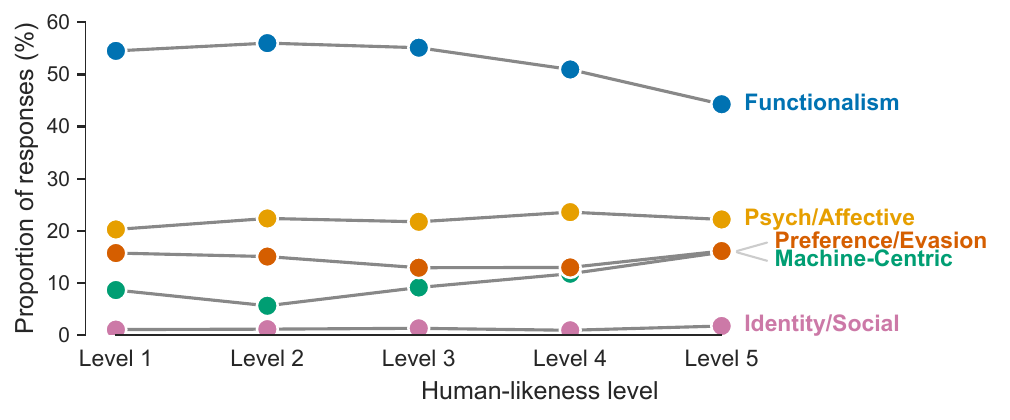}
\caption{Proportion of categories across the five human-likeness levels.}
\label{fig:category_by_hgroup}
\end{figure}

Human-likeness level significantly predicted justification patterns at both the main category ($p < .001$, $V = 0.07$) and sub-category levels ($p < .001$, $V = 0.13$).
The most prominent shift, as illustrated in Figure~\ref{fig:category_by_hgroup}, was the systematic decline of \catfunc{Functionalism} and rise of \catmachine{Machine-Centric} as human-likeness increased.
\catfunc{Functionalism} dropped from 54.5\% at Level~1 to 44.3\% at Level~5, while \catmachine{Machine-Centric} reasoning nearly doubled from 8.6\% to 15.8\%.
This trade-off was driven almost entirely by \catmachine{Neutrality} (C3.2), which rose four-fold from 2.5\% at Level~1 to 10.6\% at Level~5.

Within the \catpsych{Psych/Affective} category, the overall proportion remained stable across levels (20.2\%--23.6\%), but the internal composition shifted dramatically.
The \catpsych{Human-Likeness} sub-category (C2.3) rose thirty-fold, from 0.2\% of all responses at Level~1 to 6.4\% at Level~5, as affective reasoning shifted from task-related engagement to identity-related human resemblance with increasing anthropomorphism.

\section{DISCUSSION AND CONCLUSIONS}

Although \catfunc{Functionalism} was the primary justification for robot selection (52\%), its application was highly adaptable, consistent with functional reasoning serving as a post-hoc rationalization for underlying bias: participants justified darker robots in construction for ``hiding dirt'' and lighter robots in hospitals as appearing ``sterile.''
We might interpret this through a dual-process lens, in which unacknowledged stereotypes may drive the selection while the deliberative system generates task-based justifications. However, it is an account inferred from the alignment between written rationales and the stereotype-consistent choices documented in the companion study~\cite{he2026human} rather than from direct measures of internal motivation.
Viewed through Aversive Racism~\cite{hodson2004aversive} and the Justification-Suppression Model~\cite{crandall2003justification}, \catfunc{Functionalism} may provide a non-racial pretext for stereotype-aligned choices.
Relying solely on explicit rationales thus risks concluding that human-robot selections are objective, missing how structural bias often speaks the language of practicality.

Our findings also complicate simplistic applications of Social Identity Theory~\cite{tajfel2001integrative} and racial mirroring in human-agent interaction.
White and Asian participants showed elevated affective reasoning when selecting matching (Light or Medium) robots, yet Black participants who selected Dark robots justified their choices overwhelmingly through Functionalism (75.2\%) rather than affective ties.
This disparity suggests that identifying with a dark-skinned entity in a society with systemic bias is structurally different from identifying with a light-skinned one, perhaps because darker skin tone has been systematically stripped of ``warmth'' in cultural narratives, forcing a heavier reliance on ``competence.''

Finally, as robot human-likeness increased, participants often retreated from functional evaluations toward \catmachine{Machine-Centric} reasoning (e.g., metallic or teal colors).
This shift highlights a tension of anthropomorphic racialization: utilitarian colors read as paint on less human-like machines but as race on highly anthropomorphic forms, making functional evaluations ethically fraught.
We might interpret this retreat as a possible manifestation of racial anxiety, whereby users select non-skin-tone finishes to neutralize the robot socially, though our data document the behavioral shift rather than the psychological state directly.
This impulse parallels Color-Blind Racial Ideology~\cite{bonilla2021racism}, wherein individuals attempt to ``erase'' race in pursuit of an artificially ``neutral'' machine.

Our approach has interpretive limits.
Open-ended justifications capture the reasons participants chose to articulate, not necessarily the processes that drove their selections, so our claims about implicit bias are interpretations of written rationales rather than direct measurements of internal states.
Because socially sensitive motivations may be unstated, the patterns we report likely represent a conservative view of how bias shapes robot selection.
Methodologically, our findings derive from rendered robot illustrations and US-based participants, which may limit generalizability.
Finally, AI-assisted coding may carry the biases of its training data and is not perfectly reliable, a concern we mitigated by validating against a reconciled human consensus that yielded substantial agreement.

These findings yield three implications for social robotics.
First, practitioners must critically interrogate ``functional'' requests: requests for practical colors (e.g., brown to ``hide dirt'' in construction) may mask implicit bias, and accepting them uncritically risks hardcoding prejudice.
Second, designers face an ``anthropomorphism threshold''---assigning human skin tones to high-fidelity robots inevitably triggers racial categorization---so an explicitly non-human finish (e.g., metallic silver) is required to achieve a ``neutral'' agent.
Offering diverse skin tones (akin to digital emojis) might appear to guarantee universal affective bonding, but ``mirroring'' is an unequal experience: Black participants relied on competence and functionalism rather than affective warmth when choosing matching robots, simply offering a ``Dark mode'' robot is not enough to foster trust.
Designers may therefore need to consider the socio-historical context in which the robot operates.

Our analysis suggests that robot selection is deeply entangled with societal structures: people frequently rely on functional reasoning, and as anthropomorphism increases they retreat to non-skin-tone colors to artificially neutralize the machine.
We also challenge the assumption that racially congruent robots guarantee affective bonding, showing that systemic biases structurally alter how marginalized groups interpret these agents.

\section*{APPENDIX}
https://doi.org/10.5281/zenodo.20778157


\bibliographystyle{plain}
\bibliography{references}

\end{document}